# An Effective Fingerprint Verification Technique

Minakshi Gogoi and D K Bhattacharyya

**Abstract**—This paper presents an effective method for fingerprint verification based on a data mining technique called minutiae clustering and a graph-theoretic approach to analyze the process of fingerprint comparison to give a feature space representation of minutiae and to produce a lower bound on the number of detectably distinct fingerprints. The method also proving the invariance of each individual fingerprint by using both the topological behavior of the minutiae graph and also using a distance measure called Hausdorff distance.The method provides a graph based index generation mechanism of fingerprint biometric data. The self-organizing map neural network is also used for classifying the fingerprints.

---------------------- ◆ ----------------------

## 1 INTRODUCTION

Fingerprints are considered to be one of the most popular biometric authentication and verification measures because of their high acceptability, bility and uniqueness [1]. Here immutability refers to the persistence of the fingerprints over time whereas uniqueness is related to the individuality of ridge details across the whole fingerprint image. Automatic fingerprint tification comprises of feature extraction, fingerprint classification and fingerprint matching. The effectiveness of feature extraction depends on the quality of the images, representation of the image data, the image processing models, and the evaluation of the extracted features. At the first stage of the fingerprint classification process, the image is only represented as a matrix of grey scale sity values. Feature extraction is a process through which geometric primitives within images are isolated in order to describe the image structure, i.e. to extract important image information and to suppress redundant information that is not useful for classification and identification processes. Thus fingerprint features and their relationships provide a symbolic description of a fingerprint image. Fingerprint classification is an important step in any fingerprint identification system because it significantly reduces the time taken in identification of fingerprints especially where the accuracy and speed are critical. To reduce the search and space complexity, a systematic titioning of the database into different classes is highly essential. Fingerprint matching is done at two levels: at coarse level, fingerprints are classified into whorl, arch, tented arch, left loop, right loop or twin loop and at fine level, matching is performed by extracting minutiae i.e. ridge ending and branching points. Global ridge shape provides important clues about the global pattern configuration of a fingerprint image. One of the most desirable features of a fingerprint comparison algorithm is invariance under translation, rotation and scaling. This indicate that the problem of identifying fingerprints is of a topological nature, rather than geometrical.

In this paper we develop a graph-theoretic approach to analyze the process of fingerprint comparison for fine level matching and use it to produce a lower bound on the number of detectably distinct fingerprints. With the extracted minutiae features, a clustering algorithm is used to give a feature space representation of minutiae and a cluster graph is presented, whose topological behavior proving the invariance of each individual fingerprint

## 2 RELATED WORKS

### 2.1 Review

Based on our survey related to fingerprint classification, it has been observed that there exist different classification methods, based on the features of fingerprints. The structural features present structural classification approaches, namely syntactic pattern recognition and graph matching. Also there exist various heuristic approaches based on singularities and ridge structures. In the neural network approach, the existing applications of neural networks can also be applied. We report details of some of the classification techniques below and the comparisons of their different approaches in table.1.

**[A] Structural approach**: The structural classification approaches classify input fingerprints based on the interrelationships of low-level features.

_________________________

- *M.Gogoi is with the GIMT, Azara Ghy-17, India.*
- *D. K. Bhattacharyya is with the Department of CSE, Tezpur University, Napaam.Tezpur, Assam, India.*

TABLE 1
DIFFERENT FINGERPRINT CLASSIFICATION APPROACHES

| Approaches/ Basis | | Description |
|---|---|---|
| Structural approach | Syntactic pattern recognition | Mbayer & Fu's approach: Classes are described by a class of contextfree languages and a top-down parser is developed for classification [2]. |
| | | Chang & Fan's approach: Regular expressions for generating the ridge distribution sequences for seven fingerprint classes are formulated and a nondeterministic finite automata is constructed for classification [5]. |
| | Graph matching | Maio & Maltoni's approach: The graph represents a fingerprint's topology. For each fingerprint class a model graph is created that has a structure typical of that class [7]. |
| | | Cappelli et a's approach: A dynamic mask is defined for five fingerprint classes. For an input image, an application cost is calculated for each of the dynamic masks to create a feature vector with five elements [8]. |
| Heuristic approach | singularities | Henry's approach: Having detected a fingerprint's singularities, heuristic rules based on their number and location can be used to accurately classify fingerprints [9]. |
| | | Karu & Jain's approach: Arches have no core or delta points. Loops and tented arches both contain one core and one delta and discriminated by the symmetry of the line connecting the core and delta point[10]. |
| | global ridge structures | Chong M, Ngee T, Jun L, Gay R's approach: Based on analyzing the global geometric shape of the fingerprint. Twin loops can be recognized by the fact that they are the only global geometric shape that has two turns with opposite signs [14]. |
| | singularities and global ridge structure | Kawag e & Tojo's approach: Use singularity counts to provide a coarse classification of fingerprint images [15]. Flow-line tracing around singular regions is then used to perform more detailed classification. |
| | | Zhang et a.'s approach: Here ridge shape is incorporated with singular -rity information's approach [16],[17],[18]. The fingerprint ridges are divided into non-recurring, type-1 or type-2, based on their curvature. |
| Neural approach | KL transform | NIST for the FBI in 1990's approach: The fingerprint's directional image is registered with respect to the centre of the fingerprint image. The KL transform is used to reduce the dimensionality of the orientation field and a probabilistic neural network is used to classify the feature vector [22]. |
| | Wavelet | Net & Borges's approach: A feed-forward neural network with a single hidden layer was trained to classify feature vectors consisting of 64 wavelet coefficients [24]. |
| | SOM | Halici & Ongun's approach: The features being used for classification are the fingerprint's orientation field and some certainty measures [26]. |
| | Fuzzy network classifier | Mohamed and Nyongesa's approach: Based on the number, global direction of orientation field and relative position of core and delta points. The authors point out that noise and preprocessing errors lead to an intraclass variation among fingerprints [29]. |
| Mixing structural, statistical feature | | Nagaty's approach A three-layer feed-forward artificial neural network with six subnetworks (one for each class) is used for classification [30] |
| Cluster approach | | Wang et. al's approach: The orientation vectors is used in the area surrounding a fingerprint's core. The authors found using nine clusters had the best performance, and these clusters were found using a k -Means clustering algorithms [32]. |
| Multispace KL transform approach | | Cappelli et al's approach: One subspace is trained for each fingerprint class and fingerprints are characterized by their distances to the subspaces. MKL has a strong ability to distinguish the fingerprint classes [33]. |

a. *Syntactic pattern recognition*: In syntactic pattern recognition an analogy is drawn between the structure of the input data's features and the syntax of a language. The input data is represented by a sequence of primitives, which is considered to be a sentence of a language. Every class has an associated set of rules (or grammar) that describes how to build new sequences (sentences). Classification is performed by determining which grammar most likely produced a given input sequence.

b. *Graph matching*: In this approach two graphs are given as input, graph matching algorithms attempt to determine whether or not the graphs are isomorphic. For each fingerprint class a model graph is created that has a structure typical of that class.

[B] **Heuristic rule approach**: In this automated fingerprint classification approach, the knowledge of human experts is codified using a system of heuristic rules based on the singularity features, ridge features or a combination of singularity and ridge features.

a. *Singularity structure based approach*: Since singularities are local features they are very sensitive to noise. Having detected a fingerprint's singularities (core and delta points), heuristic rules based on their number and location can be used to accurately classify fingerprints.

b. *Global ridge structures based approach* [14]: Use the calculation of orientation fields to represent the global geometric shape of fingerprints Based on analyzing the global geometric shape of the fingerprint. Twin loops can be recognized by the fact that they are the only global geometric shape that has two turns with opposite signs.

c. *Singularities and global ridge structure based approach:* The singularities perform very poorly on noisy images. The global ridge structure feature also difficult to deal, with the large intraclass variations and small interclass variations of fingerprint classes. Some systems overcome these limitations by using both singularities and ridge structures.

[C] **Neural approach:** The research work of the applicability of neural networks to fingerprint classification began in the early 1990s and became one of the most commonly used classifiers for fingerprint classification systems. Researchers have developed different neural based classification approaches.

a. *The neural approach developed by NIST for the FBI in 1990* [22]*:* This research formed the basis for the PCASYS system (Pattern-level Classification Automation System for Fingerprints) [23]. PCASYS uses the core of loops, the upper core of whorls and a well-defined feature of arches and tented arches. The fingerprint's directional image is registered with respect to the centre of the fingerprint image. The dimensionality of the orientation field is reduced using the KL transform. Next, a probabilistic neural network (PNN) is used to classify the feature vector.

*b. The neural approach based on wavelet features:* Wavelets form the basis of the FBI's fingerprint image compression scheme [25]; however, wavelets are sensitive to rotations and translations. A feed-forward neural network with a single hidden layer was trained to classify feature vectors consisting of 64 wavelet coefficients.

*c. The neural approach based on SOM:* SOM's are based on Kohonen learning and are used for dimensionality reduction. A modified version of SOM also used that includes a certainty parameter to handle fingerprints with distorted regions. The features being used for classification are the fingerprint's orientation field and some certainty measures.

*d. The neural approach based on fuzzy-network classifier:* Combine the advantages of fuzzy logic techniques and neural networks and offer algorithms for learning and classification. The neural network is used to automatically generate fuzzy logic rules during the training period. It is based on singularity features that include the number of core and delta points, the orientation of core points, the relative position of core and delta points and the global direction of the orientation field. The authors point out that, noise and preprocessing errors lead to an intraclass variation among fingerprints. Mohamed and Nyongesa [29]

**[D] Combining structural and statistical feature:** Structural features are extracted from the orientation field using a line tracing algorithm. Prominent flow lines are represented by strings of symbols that encode information about their endpoints and curvature. A three-layer feed-forward artificial neural network with six sub networks (one for each class) is used for classification.

**[E] Clustering approach:** Uses a k -Means classifier. An unlabelled feature vector is assigned to the most common class of its three-nearest neighbours. Using clustering and three-nearest neighbours is certainly more powerful than simply using a single nearest neighbour, and it still has a low computational complexity. Clustering was performed on 500 samples, each labelled as either a whorl, left loop, right loop or arch. The features used were the orientation vectors in the area surrounding a fingerprint's core. Through experimentation, the authors found that using nine clusters had the best performance, and these clusters were found using a k -Means clustering algorithms.

**[F] Using multi-space KL transform:** The KL transform reduces the dimensionality of a feature space while minimizing the average mean-squared error. The Multispace KL (MKL) transform is a generalization of the KL transform that uses multiple subspaces for classification [33]. One subspace is trained for each fingerprint class and fingerprints are characterized by their distances to the subspaces. MKL has a strong ability to distinguish the fingerprint classes.

**[G] Support vector machines (SVMs):** SVMs are based on statistical learning theory. SVMs are binary classifiers that work by finding the optimal separating hyper plane in the feature space [36]. One advantage of SVMs is their strong ability to classify vectors with high-dimensions. SVMs are applied to the problem of fingerprint classification using the FingerCode representation of the fingerprint. SVMs are a powerful classifier and good results were presented.

**[H] Hybrid classifiers:** Uses a two-stage classifier based on FingerCodes. The algorithm first uses a k -Nearest-Neighbour classifier to determine the two most likely classes of the fingerprint. SVMs have been shown to be well suited for classifying FingerCodes [36], so by using SVMs instead of neural networks the accuracy of this system may be improved further. Classes are determined by the two most common classes of the knearest neighbours to the vector in the feature space. During the second stage, the fingerprint's class is determined by a neural network trained specifically to distinguish those two classes.

## 3 BACKGROUND OF THE WORK

Real-time image quality assessment can greatly improve the accuracy of identification system. The good quality images require minor pre-processing and enhancement. Conversely, low quality images require major preprocessing and enhancement. To test a fingerprint recognition algorithm, large databases of sample images are required to estimate error, if any occurred. But collecting large databases of fingerprint images is not a trivial task both in terms of money and time. An automatic recognition of people based on fingerprints requires that the input fingerprint be matched with a large amount of fingerprints in a database. To reduce the search space and hence computational complexity, it is desirable to classify these fingerprints in an accurate and consistent manner so that the input fingerprint be matched only with a subset of the fingerprints in the database.

### 3.1 Fingerprint Feature Extraction
**Finding of core or Reference point**
The one of the fundamental step before classification is the core point extraction, that is, the automatic detection of the core point of the fingerprint. This step is particularly important since a reference center is required in order to correctly compare two fingerprints. Automated core detection can only find the most likely center of the image without regard whether there is a meaningful core exists or not. Furthermore, the alignment according to the core point only partially remedies the misalignment of two fingerprints.

### 3.2 Fingerprint classification
Fingerprint classification is a technique to assign a fingerprint into one of the several pre-specified types. Fingerprint classification can be viewed as a coarse level match-ing of the fingerprints. An input fingerprint is first matched at a coarse level to one of the pre-specified types and then, at a finer level, it is compared to the subset of the database containing that type of fingerprints only. Fingerprints are classified into six categories: arch, tented arch, left loop, right loop and whorl and twin loop as in figure 1.

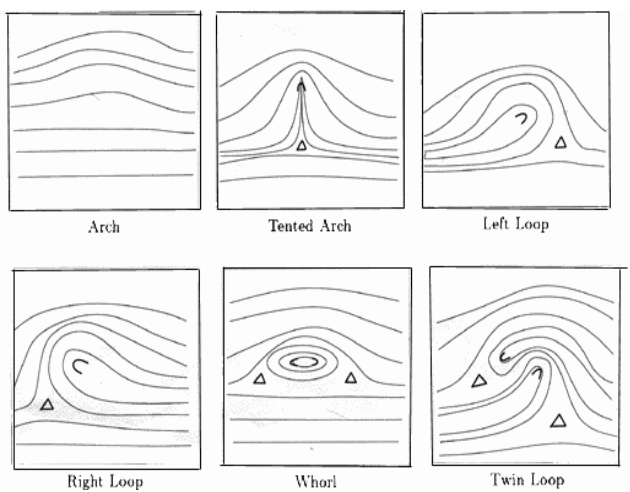

Fig.1.Fingerprint classification: arch, tented arch, left loop, right loop and whorl and twin loop

### 3.3 Fingerprint matching

Fingerprint matching is done at two levels. At coarse level, fingerprints are classified into whorl, arch, tented arch, left loop, right loop, twin loop. Coarse level classification is good only for faster detection of the class type of a given input fingerprint. At finer level, matching is performed based on the minutiae (i.e. ridge ending and branching points) information.

## 4 PROPOSED FINGERPRINT INDEX GENERATION TECHNIQUE FOR FINE LEVEL MATCHING

In our proposed system SOM based classification as depicted by [26], is used for coarse level classification whereas, a graph-theoretic approach is used to analyze the process of fingerprint comparison for finer level matching. Finer level matching is supported by extraction of minutiae i.e. ridge ending and branching points.

The dataset is used from FVC2000 DB1 and FVC2004 DB1. Also we have used a synthetic fingerprint generator SFinGe [40] to create, at zero cost, large databases of fingerprints, thus allowing recognition algorithms to be simply tested and optimized. This synthetic fingerprint generator captures the main inter-class and intra-class variations of fingerprints in nature are well enough. [41].

### 4.1. Coarse level classification
#### 4.1.1. Feature vector Generation

This phase consists of five steps. Firstly, the block directional image is carried out. Next the certainty value associated with each block is calculated .Then the segmenta-tion of the fingerprint image so that noisy and corrupted parts that do not carry valid information are deleted. Meanwhile the core point to be taken as the reference center is extracted. Then a 16x16 block directions surrounding the core are composed into the feature vector.

### I. Block Direction Estimation
The block direction estimation program operates on the gray level fingerprint image and then obtains an approximate direction for each image block with size 16x16. The technique found in [41] was used to calculate the horizontal and vertical gradients of each pixel and then combines all the gradients within the block to get an estimated direction. It has large consistent with the ridge flows of the original fingerprint image in most testing cases.

### II. Finding Certainty Values
The certainty values are generated simultaneously during the direction estimation as adopted by [26].The certainties are the magnitudes of vectors representing the flow directions of ridges in a grid for the extracted information on flow direction for the grid. The certainty vectors are of the same size as the block directional image. All the directions are then normalized into the domain from $0$ to $\pi$ , and the certain values are in the interval from 0 to 1.

### III. Segmentation
An inevitable problem after step I is that, the background noise cannot be eliminated completely, after discarding the image areas that have too small certainty values. It is needed to extract the tightly bounded fingerprint region from the image. To keep the efficiency and accuracy high, the segmentation task is accomplished by techniques like histogram equalization, image enhancement and coarse segmentation by Fourier transform, image binarization and interesting region location by morphological operations.

### IV. Finding of core or reference point
The one remaining fundamental step before classification is the core point extraction, that is, the automatic detection of the core point of the fingerprint. This step is particularly important since a reference center is required in order to correctly compare two fingerprints. An algorithm developed by [41] is used to detect the core point but with a slight difference [41], maps all the block directions to the interval from −0.5 to 0.5 and then simply regards the value 0.5 corresponds to the core.

### V. Regulate the feature vector
Preparing the feature vector is actually a trivial task. Simply we extract a 16x16 block centered at the core and then reconstruct it as a 1x256 vector. The parts of the block, in the background region or outside the image region will not affect the vector values in further process. The same operations are enforced to the certainty vector.

### 4.1.2. Algorithm for self organizing map
Assume output nodes are connected in an array and the network is fully connected (all nodes in input layer are connected to all nodes in output layer), as in figure 2.

Assume output nodes are connected in an array and the network is fully connected (all nodes in input layer are connected to all nodes in output layer), as in figure 2. In this type of neural network, the learning rate is kept large at the beginning of training epochs and decreased gradually as learning proceeds.

SOM has been used in this work as a basic classifier, where each fingerprint image is described by 256x256 pixels divided into 16x16 blocks. The orientation of each block is used as input to the neural network so the input layer consists of 256 nodes. Each image can fall into one of the five classes (right loop, left loop, whorl, arch, tented arch), so the output layer consists of five nodes. The learning rate which we used is a = 0.5 and neighborhood radius R= 0. This process is designed to facilitate the identification process whenever the system is used, thus the searching time will be reduced as the system search only in one cluster. The learning algorithm [26], [27] of such network can be described as follows:

Steps are as follows
### A. The conventional SOM
1. Construct a mxm SOM, initialize all the weights.
2. Input a fingerprint vector: $\{x_1, x_2, \ldots x_{256}\}$.
3. Find the winning node $d_{min}$:
   $$d_{min} = \min \{||\,|x\text{-}w_i|\,||\},$$
   Where $||.||$ denotes the Euclidean norm and $w_j$ is the weight vector connecting input nodes to output node j;
4. Update the weight vectors:
   $$w_{ij}(t+1) = w_{ij}(t) + L(t)[x_i(t) - w_{ij}(t)]\, N(j,t)$$
   Where $w_{ij}$ is $j^{th}$ component of the weight vector $w_j$, L(t) is the learning rate and N(j,t) is the neighborhood function.
5. Repeat 2-4 till Update is not significant

The neighborhood function is a window centered around the winning node $d_{min}$, whose radius decreases with time. In the implementation, the radiuses are simply set to decrease from the map size m to 1 during all the K runs. The learning rate function L (t) is also a decaying function. It is kept large at the beginning of training and decreased gradually as learning proceeds.

### B. Modified SOM (MSOM)

Instead of original SOM algorithms for training and classification, the modified SOM algorithm using certainty vector as parameter, as depicted at [26] is also used for classification. Here each fingerprint is associated with a certainty vector c. The steps are as follows

1. Construct a mxm SOM, initialize all weights. For reducing the variation, all weights of SOM neurons are set to zero.
2. Input a fingerprint vector:
   $X_c \{x_1, x_2, \ldots x_{256}\} = c*X + (1-c)*X_{avg}$
   where $x_{avg}$ is the vector holding the average values $x_k$ (k from 1-256) over the whole training sample space.
3. Find the winning node $d_{min}$
   where: $d_{min} = \min\{||\,|c(x\text{-}w_j)|\,||\}$
4. Update the weight vectors:
   $w_{ij}(t+1) = w_{ij}(t) + L(t)[x_i(t) - w_{ij}(t)]\, N(j,t)*c_i$
   Where $w_{ij}$ is $j^{th}$ component of the weight vector $w_j$. L(t) is the learning rate and N(j,t) is the neighbor hood function.
5. Repeat 2-4 till Update is not significant

A free SOM toolbox [31] is adopted to assist in the classification.

**Result of SOM and MSOM:** For coarse level classification we have used FVC2000, FVC2004 datasets, the result obtained is at Table 2.

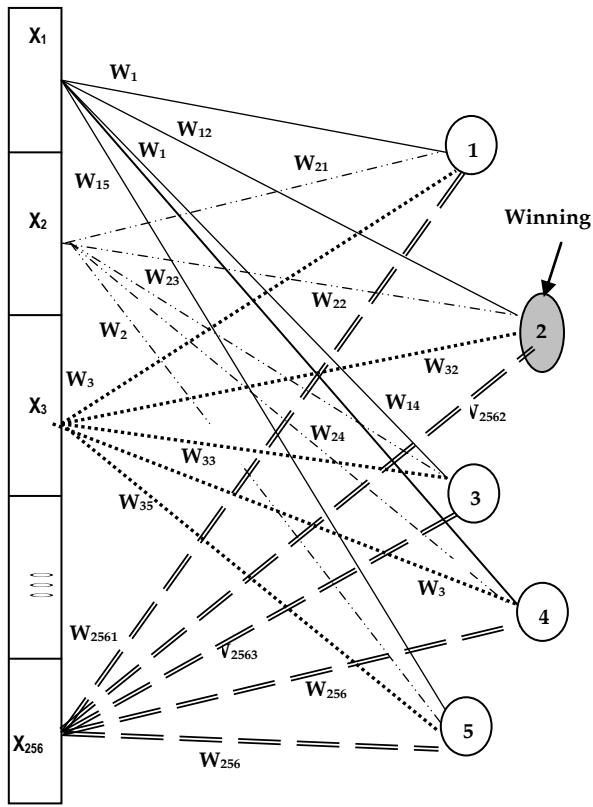

Fig.2. Basic structure for a well-trained fingerprint SOM, where winning node is 2, so the input vector X is of class 2.

*SOM construction and training*

TABLE 2
RESULTS OF CORRECT LEVEL CLASSIFICATION FOR DIFFERENT MAP SIZE

| Algorithm | Dataset size | | Map size | | |
|---|---|---|---|---|---|
| | Training | Testing | 5x5 map | 8x8 map | 10x10 map |
| SOM | 60 | 20 | 88% | 86% | 92% |
| MSOM | | | 89% | 92% | 93.47% |
| SOM | 200 | 40 | 90% | 91% | 94% |
| MSOM | | | 91% | 92.2% | 95 % |
| SOM | 600 | 40 | 91% | 93% | 94.6% |
| MSOM | | | 93.3% | 94% | 95.52% |

## 4.2 Fine level classification

### 4.2.1. Minutiae Extraction

We have used CUBS fingerprint feature extraction tool [42] for minutiae feature extraction. This tool provides a graphical user interface for minutiae feature extraction and visualization. It also allows the user to manually identify new minutiae or remove spurious ones.

### 4.2.2. Minutiae clustering for fingerprint similarity metric

After extracting the feature point set N from a fingerprint, we have clustered the feature points using K-means algorithm with variable number of clusters. The procedure minutiae_cluster( ) is given below:

**Procedure minutiae_cluster( )**

*Input*: Fingerprint minutiae $M_i$ (i = 1, 2, 3,……,n), core point $P_i(x, y)$, cluster no. m
*Output*: cluster index, $g_i$

1. For i = 1 to n do
2. Read the minutiae $M_i$
3. Call KmeansFing($M_i$, m, $\mathbf{P}_i(x, y)$) ; to perform Kmeans clustering
4. Next i.
5. Next, we report our minutiae graph generation technique.

### 4.2.3. Graph generation over clustered minutiae space

After obtaining the clustered minutiae space, cluster-graphs are generated.

**Procedure minutiae_graph()**

*Input:* cluster centroids $C_i(x,y)$ ,
*Output*: centroids distance matrix, $D_{ij}$ and graph plot.

1. For i=1 to m do
2. Read the centroids $C_i(x,y)$ ;
3. For j=1 to m do
4. Call DistMatrix($C_i,C_j$); to perform Euclidean distance
5. Next j
6. Next i
7. For i=1 to m do
8. For j=1 to m do
9. Call min_dist($D_{i,j}$)
10. draw graph $D_{i,j}$

Next, we report our graph based index generator

### 4.2.5. Graph based Index generation

From the minutiae graph, an index is generated for each fingerprint, which is unique, i.e. no two fingerprint images will have the same index. An index is generated based on four parameters:

(i ) Number of vertex
(ii) Degree of each vertex
(iii) Highest degree.
(iv) Number of vertices with same degree.

The generated index for each minutiae graph is then saved to the database.

The proposed graph based index can be found to be advantageous due to its transformation invariance; the proof of which has been stated below.

### 4.6. Proof of Transformation Invariance

To establish the proof of transformation invariance we adopt two approaches i.e. graph based and distance based. Next we report each of these approaches.

### 4.6.1. Graph based approach

We are interested in establishing that two fingerprints which are similar must be matched correctly. To do that we observe the differences comparing the minutiae cluster graphs due to small perturbations in the seed points. To accomplish this, we consider the graph obtained from the minutiae feature point clusters.

*Definition-I: Minutiae_graph*
The graph obtained from the minutiae clusters is referred as the minutiae_graph. This graph contains all the topological properties of the minutiae clusters. It tells which vertices are connected, but it does not contain any geometric measures, such as the lengths of the edges and angles they form at each vertex.

*Definition-II: Equivalent fingerprint*
Two fingerprints F and F′ are equivalent, i.e. d(F, F′)=0, if their respective minutiae_graphs are isomorphic, and the distinct, if d(F, F′)=1.

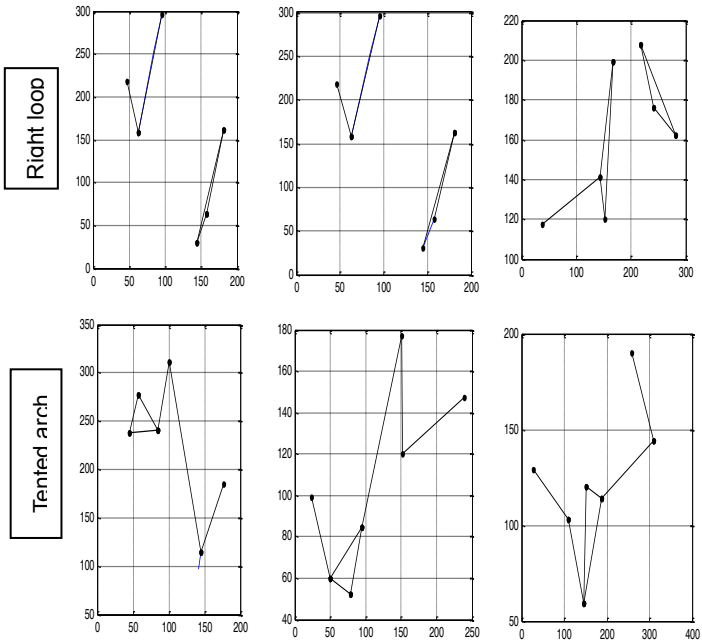

Fig. 3 Three variations of two types, viz. right loop and tented arch fingerprints graph, proofing their isomorphism under rotation and translation invariance.

Using isomorphism classes of minutiae graphs to compare fingerprints has the following advantages

1. No effect on the comparison, due to any unintentional rotation, translation, or scaling factor while recording a fingerprint.
2. A slight perturbation in the location of minutiae points will result in a topologically equivalent minutiae graph.

The cluster graphs of each individual impression are tested for isomorphism, proving their transformation invariance. The graphs obtained from two types of fingerprint, each having three impression of each; represent the following isomorphic graphs as shown in figure 3. Next we report the distance based approach.

### 4.6.2. Distance based approach

**Robust Distance Measure on Image Features:** A distance measure for the purpose of object matching should have the following properties: (1) it should have a large discriminatory power; (2) its value should increase with the amount of difference between the two objects. The operation of image matching consists of computing a measure of similarity between two images based on their features.

We have used Hausdorff distance and Modiffied Hausdorff distance (MHD) [43,44] between two sets of minutiae maps (points) associated with the fingerprint and thus proving the invariance of different fingerprint impressions.

***Hausdorff distance***: Given two finite point sets M= {$m_1$, …, $m_p$} and N ={$n_1$,…, $n_q$}, the Hausdorff distance is defined

as

$$H(M, N) = \max(h(M, N), h(N, M)) \qquad (1)$$

Where $h(M, N) = \max_{m \in} \min_{n \in} \| m - \|$

and $\| . \|$ is the underlying norm on the points of M and N. The function h(M,N) is called the directed Hausdorff distance from M to N. h(M;N) in effect ranks each point of M based on its distance to the nearest point of N and then uses the largest ranked such point as the distance . The Hausdorff distance H(M,N) is the maximum of h(M , N) and h(N , M). Thus it measures the degree of mismatch between any two shapes described by the sets M and N. Our choice of Hausdorff distance is based on its relative insensitivity to perturbations in feature points, and robustness to occasional feature detector failure or occlusion [43]

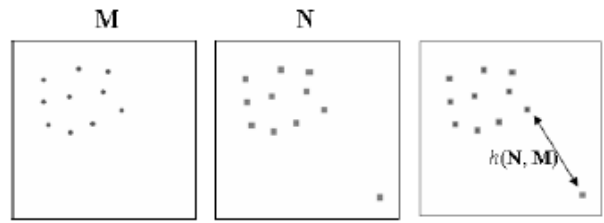

Fig. 4. The directed Hausdorff distance is large just because of a single outlier

## 5   PERFORMANCE EVALUATION

(a)  *Environment used:* The experiment was carried out on a workstation with Intel Dual-Core processor (1.86 GHz) with 1 GB of RAM. We used MATLAB 7.2 (R2006a) version in windows (64-bits) platform for the performance evaluation.

(b)  *Datasets used:* In order to evaluate the performance of the classifier, dataset is used from FVC2000 and FVC2004. The image size is of 300*300 pixels.

(c)  *Experiment Result and analysis:* From the result of our experiment we have obtained and hence proved the following two lemmas.

*Lemma1:*  Graph-based feature index for any fingerprint image remains invariant subject to translation.

*Lemma2:*  Graph-based feature index for any fingerprint image remains invariant subject to any rotational transformation.

To evaluate the performance of the reported fingerprint classifier, we have test the false accept rate (FAR) against the accuracy of the method. FAR is a measure of the fingerprints that are accepted by a certain fingerprint class, while not belonging to that particular class. An example of an event that increases the FAR of a RL (right loop type) class is a non-RL, for instance LL, fingerprint being classified as a LL fingerprint. The FAR of a particular fingerprint class is mathematically modeled as:

$$FAR = \frac{F}{S} x100$$

Where F is the total number of fingerprints that are wrongly rejected and S is the total number of fingerprints that are to be recognized. For a good fingerprint classifier, the average (AVE) FAR value should approach 0% and the value less than 20% are sufficient.

The results obtained can be depicted as in the Fig. 5. From the figure we can see that our method has both better performance and efficient, therefore is more suitable for fingerprint verification application. As the FAR value approaches to 0%, the accuracy level approaches to 100%

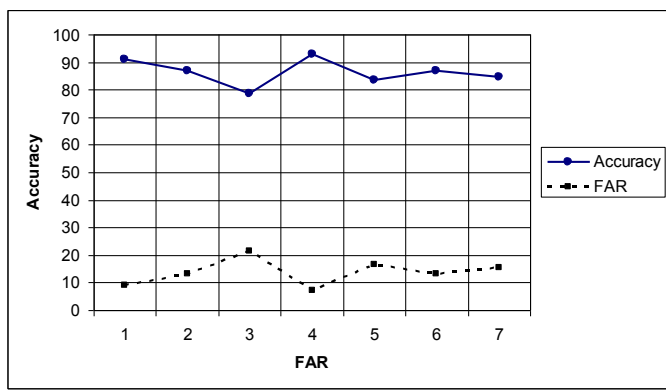

Fig. 5. FAR vs. Accuracy of MSOM based fingerprint classification

## 6 CONCLUSION AND FUTURE WORK

### 6.1 Conclusion

A limited survey on some of the popular fingerprint classification methods was carried out and found capable of identifying four or five classes with an accuracy level of (80-95) %. This paper works with a fingerprint classification method with an accuracy level of 95% for coarse level classification and presents a new approach for graph based fine level matching of fingerprints and their template generation.

Also we have reported two techniques, viz. graph based and distance based approach, for proofing robustness of various fingerprints.

### 6.2 Future work

Although fingerprint classification and matching techniques have developed drastically over times, there are scopes for developments which will make the process more efficient and accurate. A multiple SOM based approach can be used to enhance the performance.

**Minakshi Gogoi** received her MCA degree from Madurai Kamaraj University, Madurai in 2002 and received her M.Tech degree in Information Technology from Department of Computer Science and Engineering, Tezpur University in 2005. Presently she is pursuing her Ph.D. degree from the Department of Computer Science and Engineering, Tezpur University and working as a Lecturer, Department of IT, GIMT, Azara, Guwahati.Her research areas include Biometrics authentication and security, Data Mining and Image processing

**Dr. Dhruba Kr Bhattacharyya** did his PhD in Computer Science from Tezpur University in 1999. Presently he is serving as a Professor in the Computer Science & Engineering Department at Tezpur University. His research areas include Data Mining, Network Security and Content based Image Retrieval. Prof. Bhattacharyya has published more than 100 research papers in the leading Int'nl Journals and Conference Proceedings. Also, Dr Bhattacharyya written/edited 04 books. He is a Programme Committee/Advisory Body member of several Int'nl Conferences/Workshops.